\documentclass{article}
\usepackage{times}
\usepackage{graphicx}
\usepackage{subfigure} 
\usepackage{algorithm}
\usepackage{algorithmic}
\usepackage{amsfonts}
\usepackage{amsmath}
\usepackage{hyperref}
\usepackage{comment}
\usepackage{ijcai_author_kit/ijcai16}
\usepackage{xspace}

\DeclareMathOperator*{\argmin}{arg\,min}
\DeclareMathOperator*{\argmax}{arg\,max}

\usepackage{amsthm}
\newtheorem{theorem}{Theorem}

\title{Exploratory Gradient Boosting for Reinforcement Learning in
  Complex Domains}
\author{David Abel$^\ddag$ \quad Alekh Agarwal$^\dagger$ \quad Fernando
  Diaz$^\dagger$ \quad Akshay Krishnamurthy$^\dagger$ \quad Robert
  E. Schapire$^\dagger$ \\ $^\ddag$Department of Computer Science,
  Brown University, Providence RI 02912\\ $^\dagger$ Microsoft
  Research, New York NY 10011}

\usepackage{color}
\newcommand\davenote[1]{\textcolor{blue}{Dave: #1}}

\newcommand{\batchboost}{\texttt{batchboost}\xspace}
\newcommand{\booster}{\texttt{booster}\xspace}
\newcommand{\linear}{\texttt{linear}\xspace}
\newcommand{\forest}{\texttt{forest}\xspace}

\begin{document}
\maketitle

\begin{abstract}

High-dimensional observations and complex real-world dynamics present
major challenges in reinforcement learning for both function
approximation and exploration.  We address both of these challenges
with two complementary techniques: First, we develop a
gradient-boosting style, non-parametric function approximator for
learning on $Q$-function residuals. And second, we propose an
exploration strategy inspired by the principles of state abstraction
and information acquisition under uncertainty. 
We demonstrate the
empirical effectiveness of these techniques, first, as a preliminary
check, on two standard tasks (Blackjack and $n$-Chain), and then on
two much larger and more realistic tasks with high-dimensional
observation spaces.  Specifically, we introduce two benchmarks built
within the game Minecraft where the observations are pixel arrays of
the agent's visual field.  A combination of our two algorithmic
techniques performs competitively on the standard
reinforcement-learning tasks while consistently and substantially
outperforming baselines on the two tasks with high-dimensional
observation spaces.  The new function approximator, exploration
strategy, and evaluation benchmarks are each of independent interest
in the pursuit of reinforcement-learning methods that scale to
real-world domains.

\end{abstract}

\section{Introduction}
\begin{figure}[t]
\centering
\includegraphics[width=0.8\linewidth]{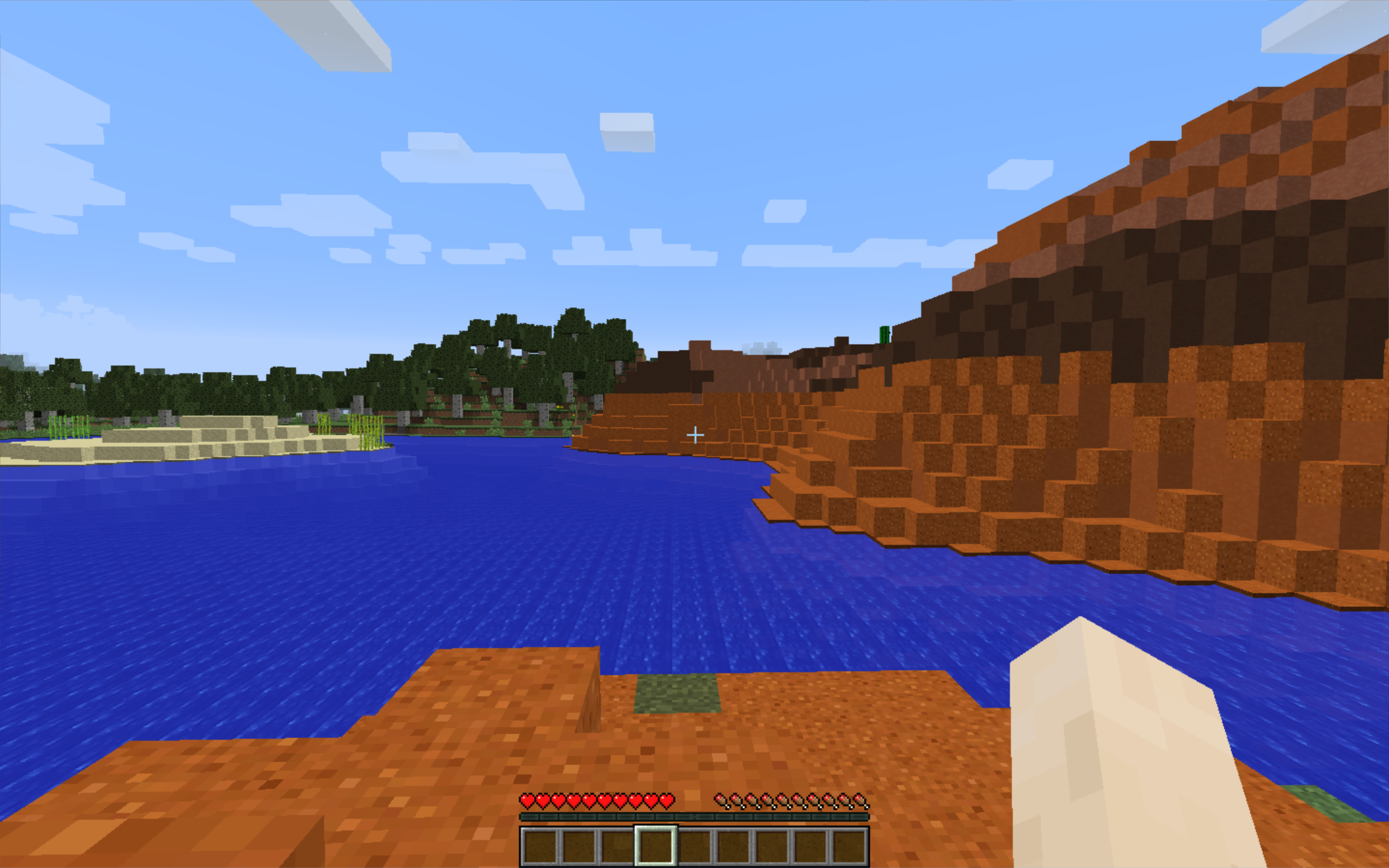}
\caption{\textbf{Visual Hill Climbing}: The agent is rewarded for
  navigating to higher terrain while receiving raw visual input.}
\label{fig:vis_hill}
\end{figure}

Many real-world domains have very large state spaces and complex
dynamics, requiring an agent to reason over extremely high-dimensional
observations.  For example, this is the case for the task in
Figure~\ref{fig:vis_hill}, in which an agent must navigate to the
highest location using only raw visual input.
Developing efficient and effective algorithms for such environments is
critically important across a variety of domains.

Even relatively straightforward tasks like the one above can cause
existing approaches to flounder; for instance, simple linear function
approximation cannot scale to visual input, while nonlinear function
approximation, such as deep Q-learning~\cite{mnih2015human}, tends to use relatively simple
exploration strategies.

In this paper, we propose two techniques for scaling reinforcement
learning to such domains: First, we present a novel non-parametric
function approximation scheme based on {\em gradient
  boosting}~\cite{friedman2001greedy,MasonBaBaFr00}, a method meant
for i.i.d.\ data, here adapted to reinforcement learning. The approach
seems to have several merits. Like the deep-learning based
methods~\cite{mnih2015human} which succeed by learning good function
approximations, it builds on a powerful learning system. Unlike the
deep-learning approaches, however, gradient boosting models are
amenable to training and prediction on a single laptop as opposed to
being reliant on GPUs. The model is naturally trained on residuals,
which was recently shown to be helpful even in the deep learning
literature~\cite{he2015deep}. Furthermore, boosting has a rich
theoretical foundation in supervised learning, the theory could
plausibly be extended to reinforcement learning settings in future
work.

As our second contribution, we give a complementary exploration
tactic, inspired by the principle of {\it information acquisition
  under uncertainty} (IAUU), that improves over $\varepsilon$-uniform
exploration by incentivizing novel action applications.  With its
extremely simple design and efficient use of data, we demonstrate how
our new algorithm combining these techniques, called {\em Generalized
  Exploratory $Q$-learning} (GEQL), can be the backbone for an agent
facing highly complex tasks with raw visual observations.  

We empirically evaluate our techniques on two standard RL domains (Blackjack~\cite{sutton1998reinforcement}
and $n$-chain~\cite{strens2000bayesian}) and on two much larger,
more realistic tasks with high-dimensional observation spaces.  Both
of the latter tasks were built within the game
Minecraft\footnote{\url{https://minecraft.net/}} where observations
are pixel arrays of the agent's visual field as in
Figure~\ref{fig:vis_hill}. The Minecraft experiments are made possible by a new Artificial Intelligence eXperimentation (AIX) platform, which we describe in detail below.
We find that on the standard tasks, our technique performs competitively, while on the two large,
high-dimensional Minecraft tasks, our method consistently and quite
substantially outperforms the baseline.


\section{Related Work}
Because the literature on reinforcement learning is so vast, we focus
only on the most related results, specifically,
on function approximation and exploration strategies.  For a more general
introduction, see~\cite{sutton1998reinforcement}.

Function approximation is an important technique for scaling
reinforcement-learning methods to complex domains.  While linear
function approximators are effective for many problems
\cite{sutton:thesis}, complex non-linear models for function
approximation often demonstrate stronger performance on many challenging
domains~\cite{anderson:thesis,tesauro:td-gammon}.  Unlike recent
approaches based on neural network architectures~\cite{mnih2015human},
we adopt gradient boosted regression trees~\cite{friedman2001greedy},
a non-parametric class of regression models with competitive
performance on supervised learning tasks.  Although similar ensemble
approaches to reinforcement learning have been applied in previous
work~\cite{marivate:ensemble-rl}, these assume a fixed set
of independently-trained agents rather than a boosting-style
ensemble.

Our work introduces the interleaving of boosting iterations and data
collection.  By its iterative nature, our approximation resembles the
offline, batch-style training of Fitted
$Q$-Iteration~\cite{ernst2005tree} in which a $Q$-learner is
iteratively fit to a fixed set of data. Our algorithm differs in that,
at each iteration, the current $Q$-function approximation guides
subsequent data collection, the results of which are used to drive the
next update of the $Q$-function. 
This adaptive data collection strategy is critical, as the exploration problem is central to reinforcement learning, and, in experiments, our interleaved method significantly outperforms Fitted $Q$-Iteration.

Our other main algorithmic innovation is a new exploration
strategy for reinforcement learning with function approximation.  Our
approach is similar to some work on state abstraction where
the learning agent constructs and uses a compact model of the
world~\cite{dietterich:maxq,Li2006}. An important difference
is that our algorithm
uses the compact model for exploration only, rather than for both
exploration and policy learning.  Consequently, model compression does
not compromise the expressivity of our learning algorithm, which can
still learn optimal behavior, in contrast with typical
state-abstraction approaches.

A number of other works propose exploration tactics for RL with
function approximation.  For example,
Oh~et~al.~\shortcite{oh2015action} train a model to
predict the future state from the current state and a proposed action,
and then use similarity of the predicted state to a memory bank to
inform exploration decisions. Another approach is to learn a dynamics
model and to then use either optimistic estimates~\cite{xie2015model} or
uncertainty~\cite{stadie2015incentivizing} in the model to provide
exploration bonuses (see also~\cite{guez2012efficient}).  Lastly,
there are some exploration strategies with theoretical guarantees for
domains with certain metric structure~\cite{kakade2003exploration},
but this structure must be known a priori, and it is unclear how to
construct such structure in general.

\newcommand{\eplen}{E}
\newcommand{\Qhat}{{\hat{Q}}}
\newcommand{\Qstar}{{Q^\star}}
\newcommand{\reals}{{\mathbb{R}}}
\newcommand{\weakclass}{{\mathcal{H}}}

\section{The GEQL Algorithm}

In this section, we present our new model-free reinforcement-learning
algorithm,
Generalized Exploratory Q-Learning
(GEQL), which includes two independent
but complementary components: a new function-approximation
scheme based on gradient boosting, and a new exploration
tactic based on model compression.

\subsection{The setting}

We consider the standard discounted, model-free reinforcement-learning
setting in which an agent interacts with an environment with the goal of
accumulating high reward. At each time step $t$, the
agent observes its state $s_t\in\mathcal{S}$, which might be represented by a
high-dimensional vector, such as the raw visual input
in Figure~\ref{fig:vis_hill}.
The agent then selects an action $a_t\in\mathcal{A}$ whose execution modifies
the state of the environment, typically by moving the agent.
Finally, the agent receives some real-valued reward $r_t$.
This process either repeats indefinitely or for a fixed number of actions.
The agent's goal is to maximize its long-term
discounted reward, 
$
\sum_{t=1}^{\infty} \gamma^{t-1} r_t
$,
where $\gamma \in (0,1)$ is a pre-specified discount factor.

This process is typically assumed to
define a Markov decision process (MDP), meaning that (1)~the next state reached
$s_{t+1}$ is a fixed stochastic function that depends only on the
previous state $s_t$ and the action $a_t$ that was executed; and
(2)~that the reward $r_t$ similarly depends only on $s_t$ and $a_t$.

For simplicity, we assume in our development that the
states are in fact fully observable.
However, in many realistic settings, what the agent observes might not
fully define the underlying state; in other words, the environment
might only be a partially observable MDP.
Nevertheless, in practice, it may often be reasonable to use
observations as if they actually were unobserved states, especially when the
observations are rich and informative.
Alternatively for this purpose, we could use a recent past window of
observations and actions, or even the entire past history.

\subsection{Boosting-based $Q$-function approximation}


\begin{algorithm}
\caption{GEQL}
\begin{flushleft}
\textsc{Input:}  Number of episodes $T$, discount factor $\gamma$, state-collapsing function $\phi$, learning rate schedule $\alpha_t$.  \\
\textsc{Output:} Policy $\pi$.
\end{flushleft}
\begin{algorithmic}[1]
\STATE $\hat{Q}(s,a)=0$ for all $s\in \mathcal{S}$, $a\in \mathcal{A}$.
\FOR{$t = 1, \ldots, T$}
    \STATE Set $s_1$ to start state
    \FOR{$i=1,\ldots,\eplen$}
        \STATE Choose $a_i$ using $\Qhat$ and IAUU exploration
        strategy 
        \STATE {Execute} $a_i$, observe $r_i$, transition to $s_{i+1}$. 
    \ENDFOR
	\STATE Set $h$ to minimize \eqref{eq:alg:q2}
            over regressors in $\mathcal{H}$
	\STATE $\hat{Q} \gets \hat{Q} + \alpha_t h$
\ENDFOR
\RETURN $\pi_{\hat{Q}}$ where $\pi_{\hat{Q}}(s) = \arg\max_{a} \Qhat(s,a) $
\end{algorithmic}
\label{alg:full}
\end{algorithm}


Our approach is based on Q-learning, a standard RL technique.
Recall that the optimal $\Qstar$ function is defined, on state-action
pair $(s,a)$, to be the expected discounted reward of a trajectory
that begins at state
$s$, with $a$ the first action taken, and all subsequent actions 
chosen optimally (to maximize the expected discounted reward).
Under general conditions, this function satisfies, for all $s,a$,
\begin{equation}   \label{eq:alg:q1}
\Qstar(s,a) = {\rm E}[r + \gamma \cdot \max_{a'} \Qstar(s',a')]
\end{equation}
where $r$ is the (random) reward and $s'$ is the (random) next state reached when
action $a$ is executed from state $s$.
Like other function-approximation schemes, ours constructs a function
$\Qhat$ that approximates $\Qstar$ by attempting to fit Eq.~\eqref{eq:alg:q1} on observed data.

As shown in Algorithm~\ref{alg:full},
we build the function $\Qhat$ iteratively in a series of episodes
using a gradient-boosting approach.
In each episode, we first use $\Qhat$ to guide the behavior of the
agent, mainly choosing actions that seem most beneficial according to
$\Qhat$, but occasionally taking exploration steps in a way that will
be described shortly.
In this way, the agent observes a series of state-action-reward tuples
$(s_i,a_i,r_i)$.
(For simplicity, we suppose each episode has fixed length $\eplen$, but
the method can easily handle variable length episodes.)

The next step is to use these observations to improve $\Qhat$.
Specifically, the algorithm fits a regressor $h$ to the
residuals between the current approximation and the
observed one-step look-ahead, as is standard for boosting methods.
That is, $h:\mathcal{S}\times\mathcal{A}\rightarrow\reals$
is chosen to (approximately) minimize
\begin{equation} \label{eq:alg:q2}
 \sum_{i=1}^{\eplen-1} [h(s_i,a_i) +\hat{Q}(s_i,a_i) - (r_i + \gamma \max_{a'} \hat{Q}(s_{i+1},a'))]^2
\end{equation}
over functions $h$ in some class $\weakclass$ of {\em weak regressors}.
As a typical example, the weak regressors might be chosen to be
regression trees, which are highly flexible,
effective, and efficiently trainable~\cite{mohan2011web}.
Once $h$ has been computed, it is added to the function approximator
$\Qhat$, thus also updating how actions will be selected on future
episodes.

This function-approximation scheme has several important advantages over existing approaches. 
First, by using a non-linear base model, such as regression trees, the agent is able to learn a complex, non-parametric approximation to the $Q^\star$ function, which is crucial for settings with high-dimensional observations.
At the same time, the training procedure is computationally efficient as updates occur in batches and trajectories can be discarded after each update.
Finally, by interleaving data collection using the learned policy induced
by $\Qhat$ with residual regression on the data so collected,
we intuitively improve the quality and informativeness of the dataset,
thus enabling the agent to more effectively and accurately approximate
the optimal $Q^\star$ function.

\subsection{The IAUU exploration strategy}
\label{sec:alg:explore}

The second novel component of the algorithm is an exploration
technique that borrows ideas from the state-abstraction literature.
The technique uses a \emph{state-collapsing function} $\phi$ which
maps each state $s$ to one of $m$ clusters,
where $m$ is relatively small.  In GEQL, this function
is trained by clustering a large dataset of states, for instance, using the
$k$-means algorithm (as in all our experiments), and associating each state $s$
with the nearest cluster center.  Ideally, an optimality-preserving
compression would be used, such as those characterized
by Li \emph{et al.},~\shortcite{Li2006},
but such abstractions are not always easily computable.

The main idea of our technique is to keep track of how often each
action has been taken from states in each of the clusters, and to
choose exploratory steps that encourage the choice of actions that
were taken less often in the current cluster.
Thus, on each episode, for each of the $m$ clusters $c$, and for each
action $a$, we maintain a count $M(c,a)$ of how often action $a$ was
taken from states $s$ in cluster $c$, i.e., for which $\phi(s)=c$.
For each state $s$, this table induces a Gibbs distribution
over actions defined by $p(a|s) \propto \exp\{-\rho M(\phi(s),a)\}$
where $\rho>0$ is a temperature parameter that controls the uniformity
of this distribution.  In concert with the current function
approximator $\hat{Q}$, we use this distribution to define a
randomized choice of actions at each step of the episode.
Specifically, when in the current state $s$, with probability
$1-\varepsilon$, we choose to act greedily, selecting the action $a$ that
maximizes $\Qhat(s,a)$; and with probability $\varepsilon$, we take
an exploration step, sampling $a$ from
$p(a | s)$.
We call this exploration strategy \emph{Information Acquisition Under Uncertainty} or IAUU. 

This strategy shares some of the benefits of the work on state abstraction without suffering from the drawbacks. 
The main advantage is that the state-collapsing function promotes applying infrequently-taken actions, thereby encouraging the agent to visit new regions of the state space.
However, in contrast with the state-abstraction literature, the method remains robust to misspecification of the state-collapsing function since we use it only for exploration so that it does not compromise the agent's ability to learn optimal behavior.
Finally, IAUU exploration adds minimal computational overhead as the space needed is $O(m|\mathcal{A}|)$ and the additional running time per action remains $O(|\mathcal{A}|)$. 

The IAUU exploration tactic is not attached to the function
approximation scheme, so it can also be used with $Q$-learning in the
tabular setting, i.e., when $\Qhat$ is maintained explicitly as a table.
In this case, there are a number of modifications to the exploration
strategy under which the convergence properties of $Q$-learning can be retained. 
One option is to modify the exploration distribution $p(a|s)$ to mix
in a vanishingly small amount of the uniform distribution.
Another option is to only count exploration steps when updating the state-visitation table $M$.
In both cases, each action is taken from each state infinitely often, and this property suffices to ensure convergence of $Q$-learning in the tabular setting. 

\section{Experiments on Standard Benchmarks}

This section details our evaluation on the two standard reinforcement learning benchmarks of Blackjack and $n$-Chain.

\subsection{Blackjack}
We conducted experiments on the Blackjack domain, implemented exactly as defined in Section 5.1 of Sutton and Barto~\shortcite{sutton1998reinforcement}. 
In this experiment, we test the hypothesis that GEQL will improve over standard baselines even in fairly simple domains without high-dimensional visual input. In particular, Blackjack is fully observable, has low noise, low dimension, short episodes, and a small action space. 

\textbf{Algorithms and Parameters:} For GEQL, we used depth-2
regression trees as the \emph{weak regressors}, fit using Python's
\emph{scikit-learn} package~\cite{buitinck2013api}.  To test and
isolate the effectiveness of our incremental boosting approach
(henceforth \booster), we compared GEQL against three baseline
function approximators:
\begin{enumerate} 
\item $Q$-learning with a linear approximator (\linear)
\item $Q$-learning with a Batch Boosted Regression approximator, similar to Fitted Q-Iteration (\batchboost)
\item $Q$-learning with a Batch Random Forest Regression approximator (\forest).
\end{enumerate}
Each approximator used the same set of features for all
experiments. The two batch-based regression approaches were trained
after every 50 episodes. Similar to GEQL, we set the depth of the
regression trees for each of the batch approaches to two. The
\batchboost and \forest approximators used the same number of total
trees as GEQL, but were trained in batch. That is, if we run for 500
episodes, then each tree based approach gets a total of 500 trees. For
\batchboost and \forest, all the 500 trees are retrained on 50
episodes worth of data, while \booster adds a new tree every episode.

We ran all function approximators
with $\epsilon$-uniform exploration and with IAUU exploration.  Across
all experiments, we set $\epsilon_{0}=0.4$, $\alpha_{0}=0.15$,
$\gamma=0.95$, and decayed so that, during episode $t$, $\epsilon_t =
{\epsilon_{0}}/{(1 + 0.04 t)}$, and $\alpha_t = {\alpha_{0}}/{(1 +
  0.04 t)}$. The state clustering function used for IAUU was learned for
  each individual task by randomly sampling states via a random policy.
\begin{figure}[t]
\includegraphics[width=\linewidth]{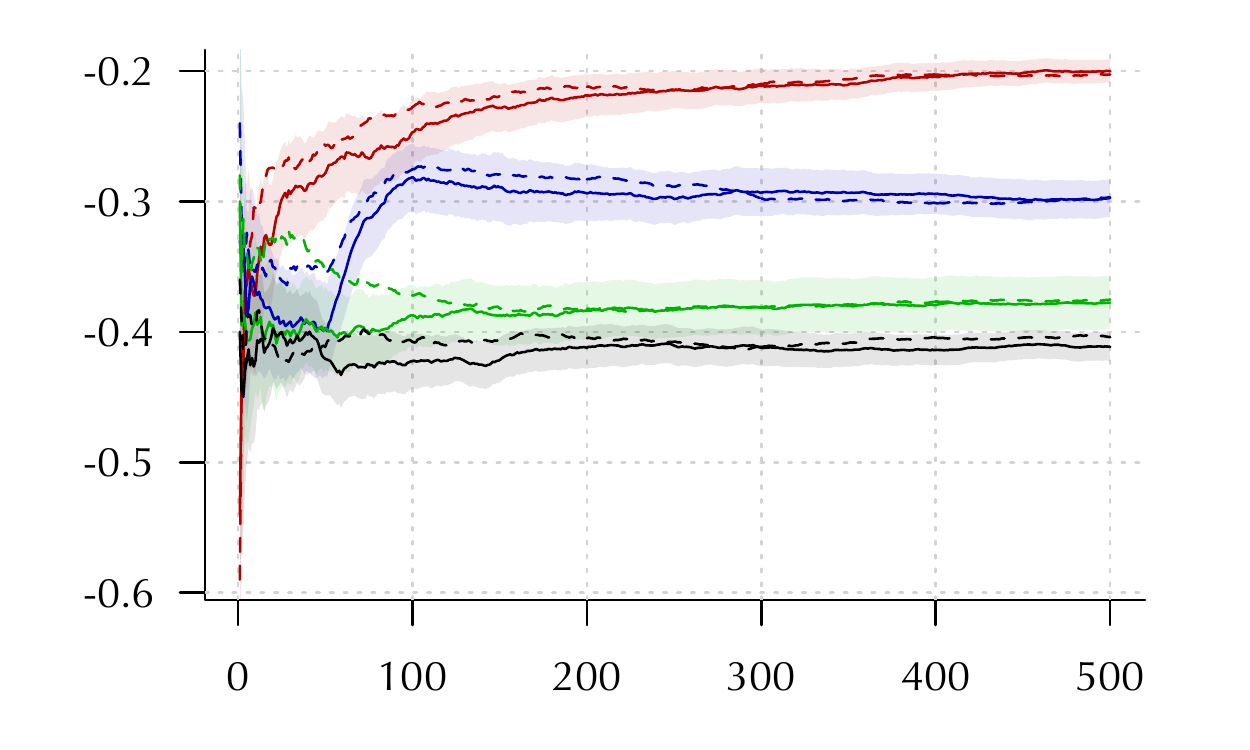}
\vspace{-0.4in}
\caption{\textbf{Blackjack}: Running average per-episode reward with error band denoting $2$ standard errors for IAUU runs (\textbf{Legend:} red: \booster, blue:\forest, green: \batchboost, black: \linear; solid: IAUU, dashed: uniform).}
\label{fig:blackjack_results}
\end{figure}

\textbf{Results:} Figure~\ref{fig:blackjack_results} shows results from 100 trials run for 500 episodes each. The results indicate that during these 500 episodes, very minimal learning occurred for the linear approximator, while the gradient-boosting approximator was able to learn to play far more effectively (yielding a statistically significant improvement in performance). The two batch approximators demonstrated some learning, though the gradient-boosting approximator far outperformed both of them. However, the exploration tactic had a negligible effect in this domain, likely due to the small action space (two actions), and short episodes (typically between one and two actions per episode). The brevity of each episode may also explain why the linear approximator learned so little in this many episodes.

\subsection{$n$-Chain}

We conducted experiments on the $n$-Chain domain from Strens~\shortcite{strens2000bayesian}.
In this domain, there are $n$ states numbered $0,\ldots,n-1$, each with two actions available: applying the {\it forward} action in state $i$ advances the agent to state $i+1$, while the {\it return} action moves the agent to state $0$. 
Applications of the {\it forward} action provide zero reward in all states except transitions to state $n-1$, which provides reward $100$.
Applications of the {\it return} action receive reward $2$ for all transitions.
Both actions are also stochastic, so that with probability $0.2$, they have the opposite effect. 
This task poses a challenging exploration problem since the agent must avoid greedily taking the {\it return} action to learn optimal behavior by exploring all the way to the last state in the chain.
We used $n=5$, which is a typical setting for this task.

\textbf{Algorithms:} As this is a tabular problem, we evaluated only tabular methods. 
We used a tabular $Q$-learner with uniform exploration, IAUU exploration, and the $\textsc{RMax}$ algorithm~\cite{brafman2003r}, which is a model based algorithm with strong sample complexity guarantees.

\begin{figure}[t]
\includegraphics[width=\linewidth]{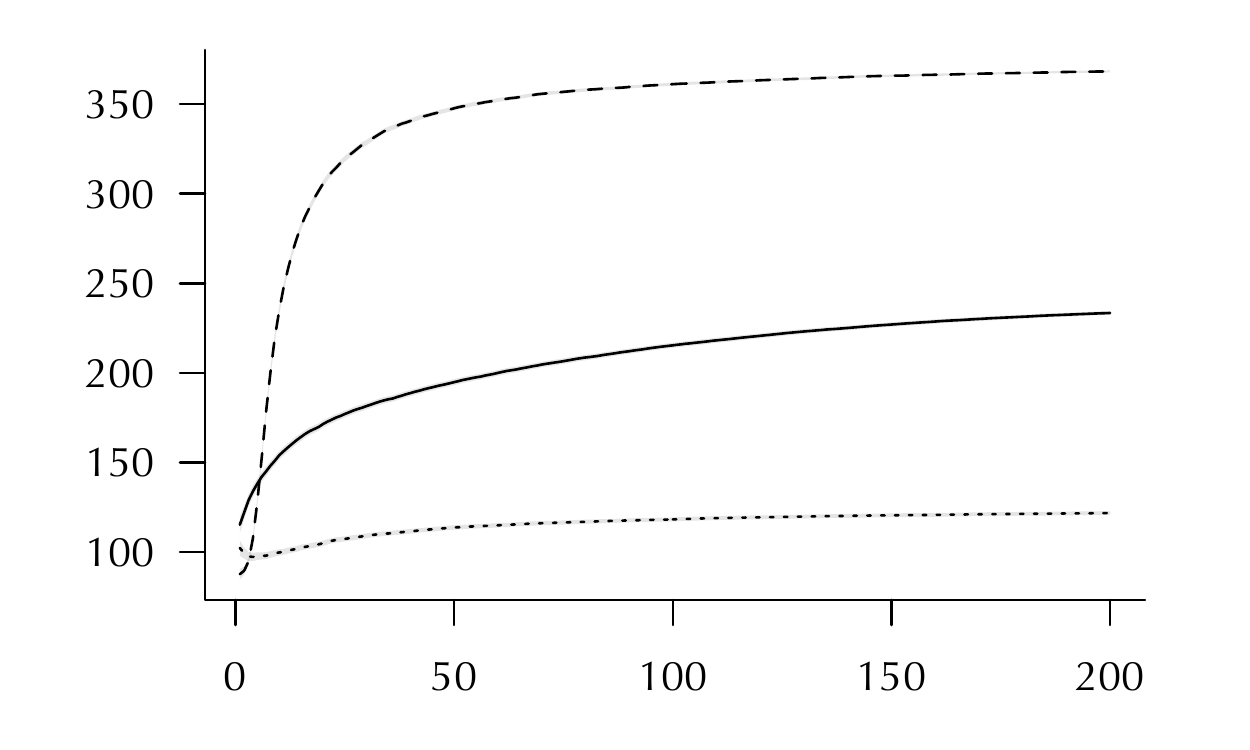}
\vspace{-0.4in}
\caption{\textbf{$n$-Chain}: Running average per-episode reward (\textbf{Legend:} IAUU: solid, $\textsc{RMax}$: dashed, $Q$-learning: dotted)}
\label{fig:nchain_results}
\end{figure}

\textbf{Results:} Figure~\ref{fig:nchain_results} displays results from 5000 trials. Unsurprisingly, $\textsc{RMax}$ significantly outperformed the two tabular $Q$-learning strategies used here since it is designed to seek out all state-action applications helping it to quickly discover the high reward at the end of the chain.
$Q$-Learning with $\epsilon$-uniform exploration, on the other hand, will only discover the high reward in the last state of the chain with exponentially small probability since the agent will favor the greedy action and must repeatedly explore to advance down the chain.
The IAUU exploration is between these two extremes; it generalizes to extremely large state spaces unlike $\textsc{RMax}$, but provides more effective exploration than $\epsilon$-uniform.

\section{Experiments on Visual Domains}

This section describes our empirical evaluation on two highly challenging problems in which the agent must reason over raw RGB images.
We used the Minecraft game platform to provide environments for the two tasks.
Minecraft is a 3D blocks world in which the player can place blocks, destroy blocks, craft objects, and navigate the terrain. The size of the underlying state space grows exponentially with the number of blocks allowed in the world, which is typically on the order of millions, making planning and learning infeasible for tabular approaches. 
Moreover, there are day and night cycles and weather cycles that dramatically alter the visual world, as well as animals that roam the world, and underwater environments.
The size of the state space and these complex visual elements pose significant challenges to learning from raw visual inputs.
Minecraft has previously been used in planning research~\cite{abel:goal-based-action-priors}, and has been advocated as a general artificial intelligence experimentation platform~\cite{aluru:minecraft-platform}.

Our experimental results here are enabled by a recently developed
Artificial Intelligence eXperimentation (AIX) platform, designed for
experimentation within Minecraft.
AIX provides a flexible and easy-to-use API to the Minecraft engine
that allows for full control of an agent, including action execution
and perception, as well as precise design of the Minecraft world the
agent operates in (i.e. specific block placement, day and night
cycles, etc.). For the Visual Grid World task, we hand crafted a grid
world inspired by the classical reinforcement learning
task~\cite{russell1995modern}, while the Visual Hill Climbing task was
built by Minecraft's random world generator. With AIX running,
Minecraft runs around 30 frames per second, though our agents only
execute around 2 actions per second.

\subsection{Implementation Details}
\textbf{Visual System:}
In a rich 3D world such as Minecraft, reinforcement learning agents require additional preprocessing of raw RGB observation.
We employ classical computer vision techniques for preprocessing of the raw visual images from the Minecraft game. 
The input to the visual pipeline is a $320 \times 240$ image of the Minecraft agent's view, with all distracting UI elements (such as the toolbar) removed.

We use data from a five-minute random exploration of the Minecraft
world (in which the agent takes a random
action every time step). Specifically, for every
20th frame received from the game engine, the agent performs SURF
key-point detection~\cite{bay2008speeded}, and stores the set of key
points in a dataset. After five minutes, the agent performs $k$-means
clustering on the space of all key points to reduce the dimensionality
of the key-point space of interest. This training is done offline before
experiments are conducted. The same visual system is used for all
algorithms. The system is trained separately for each task. 


During the task, for each new frame the agent receives, it partitions
the frame into a $3\times 3$ grid. For each partition, the agent finds
the key point that is most similar to each of the key-point centers,
and computes its distance. This distance is used as a feature for
that cell. The ultimate feature vector is the concatenation of each
partition's key-point distances (so if $k = 10$, there will be $10$
features per $3 \times 3$ partitions for a total of $90$ features plus
a bias term).

\textbf{Occupancy Grid:}
Since the RGB image available to the vision system is based on the agent's first-person perspective (see Figure~\ref{fig:vis_hill}), the agent's immediate surroundings are partially occluded.
As immediate surroundings are crucial to decision making, we augment the agent with a $4 \times 3 \times 3$ 
\emph{occupancy grid} of the cells the agent is touching. This occupancy grid contains a $1$ if the corresponding cell adjacent to the agent contains a block that is solid (e.g. dirt, grass, stone, etc.) or water, and a $0$ otherwise.
These binary features, along with the key point distances from the vision system, comprise the state feature vector available to the agent. 

\textbf{State-Collapsing Function:} For the state-collapsing function
$\phi$ (see Section~\ref{sec:alg:explore}), we train another $k$-means
instance that maps a given state object to a lower dimension. We let
the Minecraft agent explore for another five minutes, saving every
20th frame. For each saved frame, the agent computes the feature
vector as described above and concatenates the occupancy grid of the
agent's surrounding cells, storing these vectors in a data set. After
five minutes, the agent performs $k$-means on the data set of features
to reduce the feature space to a lower dimension. The training is also
done offline before experiments are conducted, and all IAUU algorithms
use the same state-collapsing function for each task.

During learning, IAUU agents evaluate the state-collapsing function by mapping the current state's feature vector to the nearest cluster center in this $k$-means instance. 

\subsection{Visual Grid World}
\begin{figure}[t]
\centering
\includegraphics[width=0.8\linewidth]{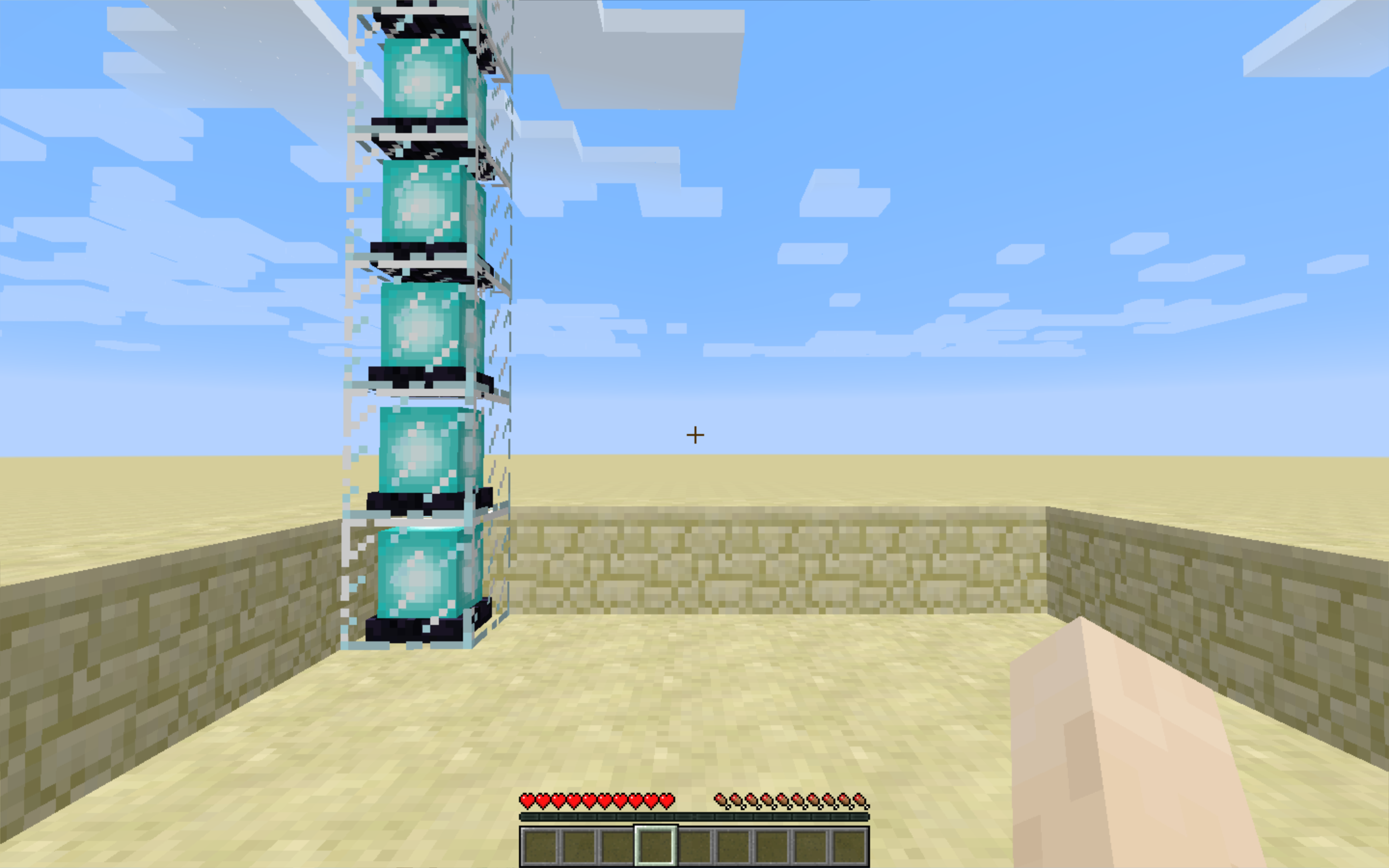}
\caption{\textbf{Visual Grid World}: The agent is rewarded for navigating to the blue pillar while receiving raw visual input.}
\label{fig:vis_grid}
\end{figure}

The first task we consider is the \emph{Visual Grid World} task.
Here, the environment consists of a $6\times 6$ grid.  The agent always starts at $(0,0)$ and must navigate to $(5,5)$ using only movements North, East, South, and West.  There is no rotation (i.e. the agent is always facing North).  At each state, the agent observes the raw bitmap image of the agent's view (Figure~\ref{fig:vis_grid}), which is preprocessed using the vision system and augmented with the occupancy grid. The reward function is the negation of the agent's Euclidean distance away from the goal.  For example, if the agent is a distance of 5 from the goal, the agent receives $-5$ reward.  All transitions are deterministic.  The optimal policy will achieve a reward of roughly $-31$ for directly proceeding to the goal.

\textbf{Algorithms and Parameters:} As in Blackjack, we used four
function-approximation schemes (\linear, \batchboost, \forest, and our
interleaved gradient-boosting approach denoted as \booster), and two
exploration strategies ($\epsilon$-uniform and IAUU).  Regression
problems are solved after each episode for the linear and gradient
approximators using data from only the most recent episode.  The
\batchboost and \forest approximators are completely retrained every
five episodes, using only the most recent five episodes of data.  As
before, the depth and number of trees in all tree-based methods is
set to two and the total number of episodes (100).  Other parameters
are set as in the Blackjack experiment.

\begin{figure}[t]
\includegraphics[width=\linewidth]{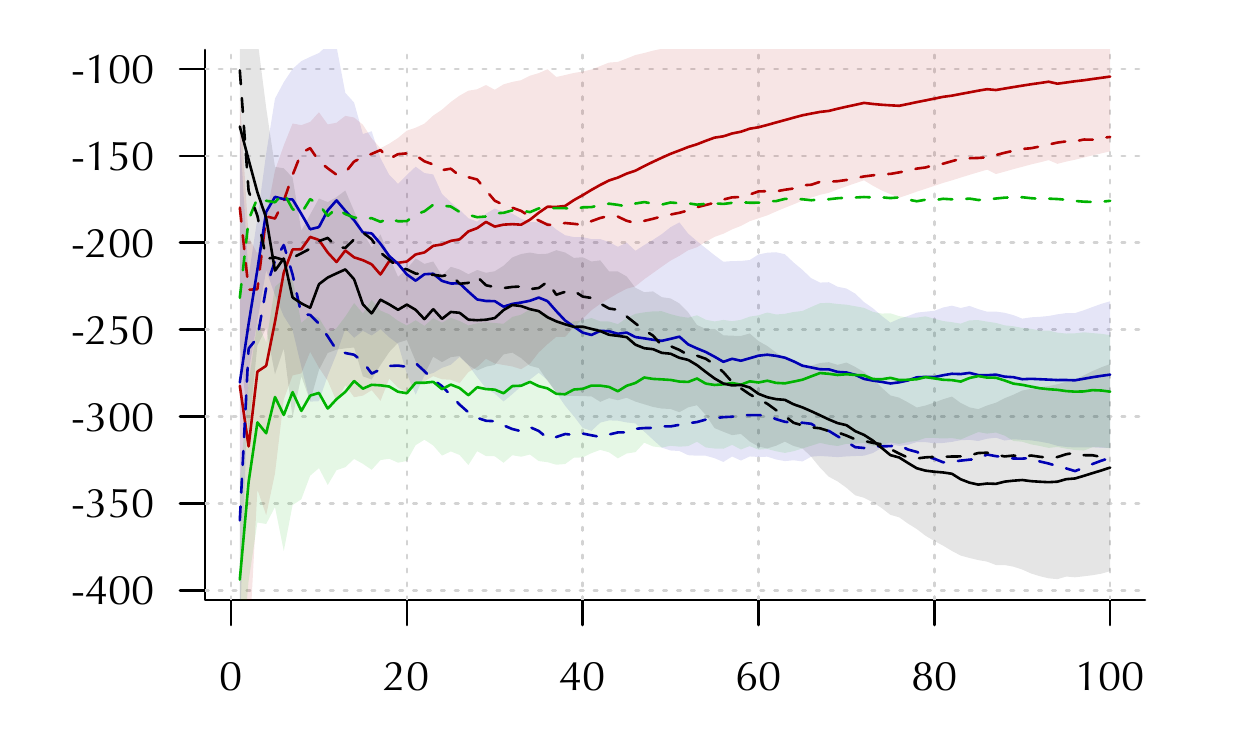}
\vspace{-0.2in}
\caption{\textbf{Visual Grid World}: Running average per-episode reward with error bands on IAUU versions, denoting $2$ standard errors (\textbf{Legend:} red: \booster, blue:\forest, green: \batchboost, black: \linear; solid: IAUU, dashed: uniform).}
\label{fig:visual_grid_results}
\end{figure}

\textbf{Results:} Figure~\ref{fig:visual_grid_results} shows results
from five trials for 100 episodes, where each episode was at most 40
seconds.  Results demonstrate that the gradient-boosting approximator
led to dramatically faster learning on this task than the linear
approximator (statistically significant at the $0.05$ level).
\booster also outperforms the two batch approximators with similar
statistical significance, apart from the \batchboost baseline with
uniform exploration (which still underperforms, though
insignificantly).
While the combination of gradient boosting and IAUU exploration has
the best average performance, the improvement over gradient boosting
with $\epsilon$-uniform is not statistically significant.
Due to the speed of the Minecraft engine, scaling these experiments is
challenging.  Nevertheless, these results clearly show that GEQL
(\booster with IAUU exploration) is a major improvement over all
previous baselines for this challenging task.

Since we know the reward of the optimal policy in this case, we also
checked the reward for the policy learned by \booster at the end of
100 episodes. We found that the average rewards of \booster with IAUU
and uniform explorations were $-34.22$ and $-82.77$ respectively. Note
that the optimal policy gets a reward close to $-31$, while the best
baseline of \batchboost with uniform exploration picks up a reward
around $-119.5$ in this task. Hence, we can conclude that GEQL learns
a significantly better policy than other baselines on the task.

\subsection{Visual Hill Climbing}
The second task, which we call \emph{Visual Hill Climbing}, is
especially difficult. At its core, it is a variant of non-convex
optimization in a 3D environment. As with Visual Grid
World, the state is the preprocessed raw RGB bitmap image of
the agent's view along with the occupancy grid. The agent must climb
the highest hill it can find by navigating a highly complex 
3D world that includes animals, lakes, rivers, trees,
caverns, clouds, pits, and a variety of different terrain types. An
example snapshot of the agent's task is pictured
in~Figure~\ref{fig:vis_hill}. This is an especially challenging
exploration problem as the agent's view is restricted to a finite
horizon, and the agent can only see in one direction at a time. Large
hills may also be partially occluded by trees, animals, the agent's
own arm, and other hills. Scaling some hills involves steps with jumps
larger than agent can make in one action, meaning the agent cannot
scale the hill even if it gets there.

The agent may move forward in the direction it is facing, turn left 90 degrees, turn right 90 degrees, jump up two units, or perform a combination that jumps up two units and then moves forward.
The agent receives $+N$ reward for increasing its elevation by $N$ units,  $-N$ reward for decreasing its elevation by $N$ units, and a small reward proportional to its current height relative to ``sea level''.  So, if the agent's initial elevation is $0$, then if the agent reaches an elevation of $10$, it will receive an additional $\frac{10}{1000}$ reward.  All transitions are deterministic, but the state is only partially observable, so repeated application of an action at a particular observation may result in dramatically different future observations.

\textbf{Algorithms and Parameters:} We used the same algorithms and
parameter settings as for the Visual Grid World.

\begin{figure}[t]
\includegraphics[width=\linewidth]{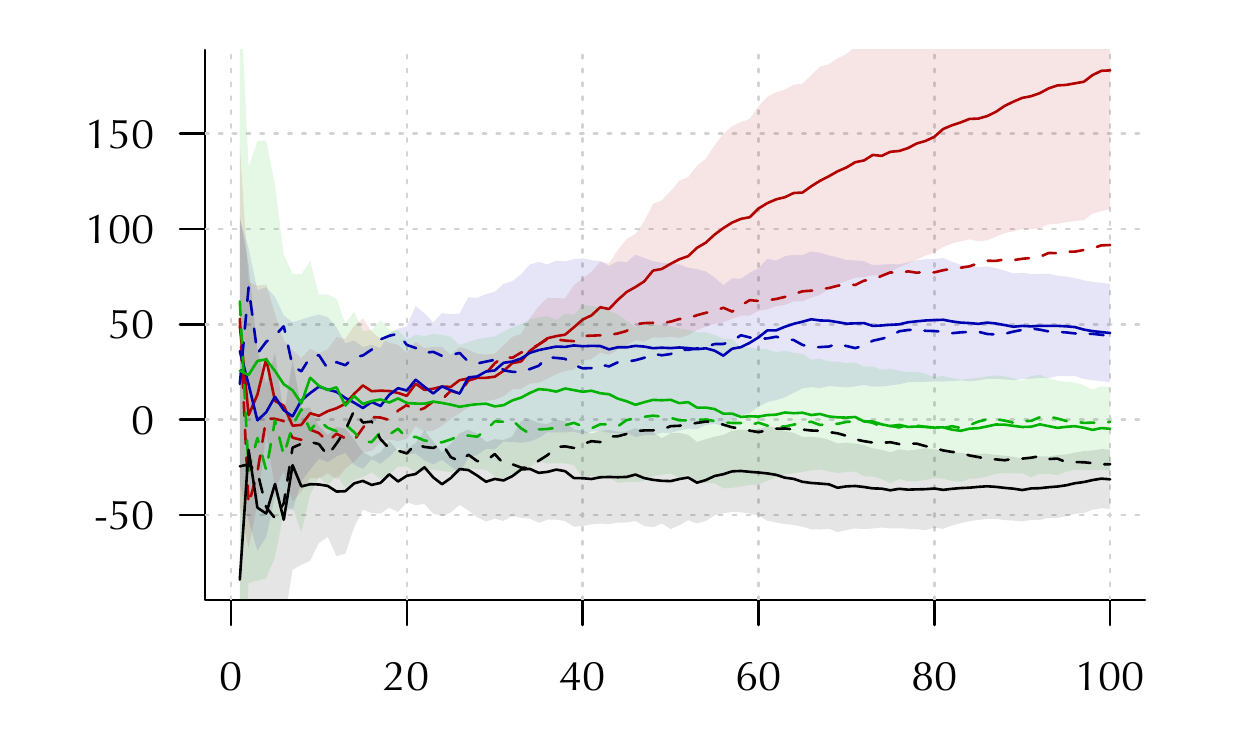}
\vspace{-0.2in}
\caption{\textbf{Visual Hill Climbing}: Running average per-episode reward  with error bands on IAUU versions, denoting $2$ standard errors (\textbf{Legend:} red: \booster, blue:\forest, green: \batchboost, black: \linear; solid: IAUU, dashed: uniform).}
\label{fig:visual_hill_climb_results}
\end{figure}

\textbf{Results:} Figure~\ref{fig:visual_hill_climb_results} displays results from ten trials for 100 episodes, where each episode was exactly 40 seconds.
Again, results indicate that the gradient booster is able to learn a far better policy than any of the other approximators, and that the IAUU exploration tactic helps.
Indeed, only for the gradient booster does non-negligible learning occur; given how complex this domain is, it is extremely promising that we can learn a reasonable policy at all (i.e. that the agent is able to learn a policy that leads to positive reward, suggesting that the agent is climbing a substantial amount).

\begin{figure*}
  \includegraphics[width=\textwidth]{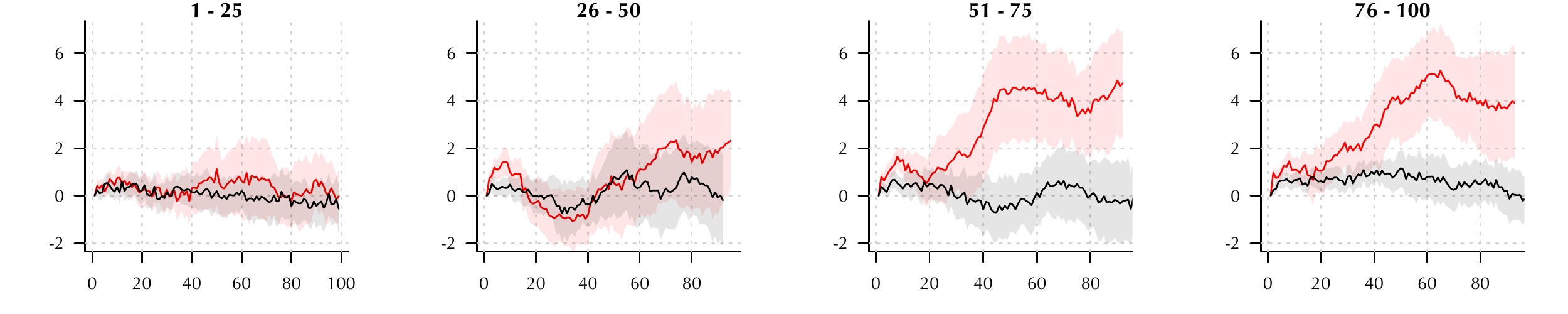}
  \vspace{-0.2in}
  \caption{\textbf{Visual Hill Climbing Elevation Profile:}  Elevation throughout an episode averaged over the episodes indicated in the range (i.e. the first, second, third, and fourth 25 episodes) for \booster (red) and \forest (black), both with IAUU exploration. All elevations are with respect to the starting elevation.  The elevation of an effective agent should go up.}
  \label{fig:elevation}
\end{figure*}

To visualize the performance of our agent better, we plot the
elevation profile of the policy learned by \booster with IAUU
exploration over time in Figure~\ref{fig:elevation}. We notice that
while the agent barely increases its elevation in the initial quarter
of the episodes, it is reliably reaching much better altitudes in the
last quarter indicating that it does identify the key to succeeding at
this task.

\section{Discussion}

In this paper, we have described novel approaches to function
approximation as well as exploration in reinforcement learning, and
evaluated it on challenging tasks implemented within Minecraft. The
encouraging performance of our methods suggests several other exciting
avenues for future research. Empirically, the performance of gradient
boosting coupled with its favorable computational properties appears
very promising, and it would be interesting to compare with
more computationally-intensive deep-learning based approaches in future work. Extending the existing
theory of gradient boosting from supervised to reinforcement
learning is also a natural question.

In terms of exploration, while IAUU certainly improves over
$\epsilon$-uniform, it is still limited by the state-collapsing
function used, and can be suboptimal if the least-frequent action also
happens to be bad. It remains challenging to find better
alternatives that are tractable for reinforcement learning with
real-time decisions and high-dimensional observations.

Finally, Minecraft provides an attractive framework to develop visual versions of standard
RL tasks. We show two examples here,
but the opportunity to translate other tasks that stress and
highlight various learning abilities of an agent, as well as
understand the gulf to human performance 
is a very exciting proposition for future work.

\textbf{Acknowledgement:} We would like to thank Katja Hoffman, Matthew Johnson, David Bignell, Tim Hutton, and other members of the AIX platform development team, without whose support this work would not be possible.


\bibliographystyle{named}
\bibliography{arxiv_geql}

\end{document}